\documentclass[11pt]{article}

\usepackage{times}
\usepackage{amsmath, amssymb}
\usepackage{graphicx}
\usepackage{url}
\usepackage{hyperref}
\usepackage{natbib}   
\usepackage{algorithm}
\usepackage{algorithmic}
\usepackage[utf8]{inputenc}
\usepackage[affil-it]{authblk} 

\usepackage[symbol*]{footmisc} 

\usepackage[utf8]{inputenc}
\usepackage[T5]{fontenc}
\makeatletter
\def\@maketitle{%
  \null
  \hrule height 1pt
  \vskip 1em%
  \begin{center}%
    {\LARGE \bfseries \@title \par}%
    \vskip 1em%
  \end{center}%
  \hrule height 1pt
  \vskip 1.5em%
  \begin{center}%
    {\large \bfseries \@author \par}%
  \end{center}%
  \par
  \vskip 1.5em}
\makeatother

\title{xAI-CV: An Overview of Explainable Artificial Intelligence in Computer Vision}
\author[a,b]{Nguyen Van Tu}
\author[a,b]{Pham Nguyen Hai Long}
\author[a,b,*]{Vo Hoai Viet}

\affil[a]{Faculty of Information Technology, University of Science, Ho Chi Minh City, Vietnam}
\affil[b]{Vietnam National University, Ho Chi Minh City, Vietnam}

\date{} 

\usepackage[a4paper, left=2.5cm, right=2.5cm, top=2.5cm]{geometry}

\begin{document}

\maketitle

\footnotetext[1]{Corresponding author. Email: vhviet@fit.hcmus.edu.vn}

\begin{abstract}
Deep learning has become the de facto standard and dominant paradigm in image analysis tasks, achieving state-of-the-art performance. However, this approach often results in “black-box” models, whose decision-making processes are difficult to interpret, raising concerns about reliability in critical applications. To address this challenge and provide human a method to understand how AI model process and make decision, the field of xAI has emerged. This paper surveys four representative approaches in xAI for visual perception tasks: (i) Saliency Maps, (ii) Concept Bottleneck Models (CBM), (iii) Prototype-based methods, and (iv) Hybrid approaches. We analyze their underlying mechanisms, strengths and limitations, as well as evaluation metrics, thereby providing a comprehensive overview to guide future research and applications.

\end{abstract}

\section{Introduction}
Over the past decade, deep learning models have revolutionized the field of computer vision, achieving outstanding performance in a wide range of tasks from image classification to medical diagnosis. However, this power often comes at a significant cost: these models operate as complex “black boxes”, making their decision-making processes difficult to understand. This lack of transparency poses significant barriers to trust, fairness, and acceptance in high-stakes applications such as autonomous vehicles or clinical medicine. To address this challenge, the field of xAI emerged with the core objective of clarifying the inner workings of models, building trust, and enabling effective interaction between humans and machines.

\begin{figure}[H]
\centering
\includegraphics[width=0.8\linewidth]{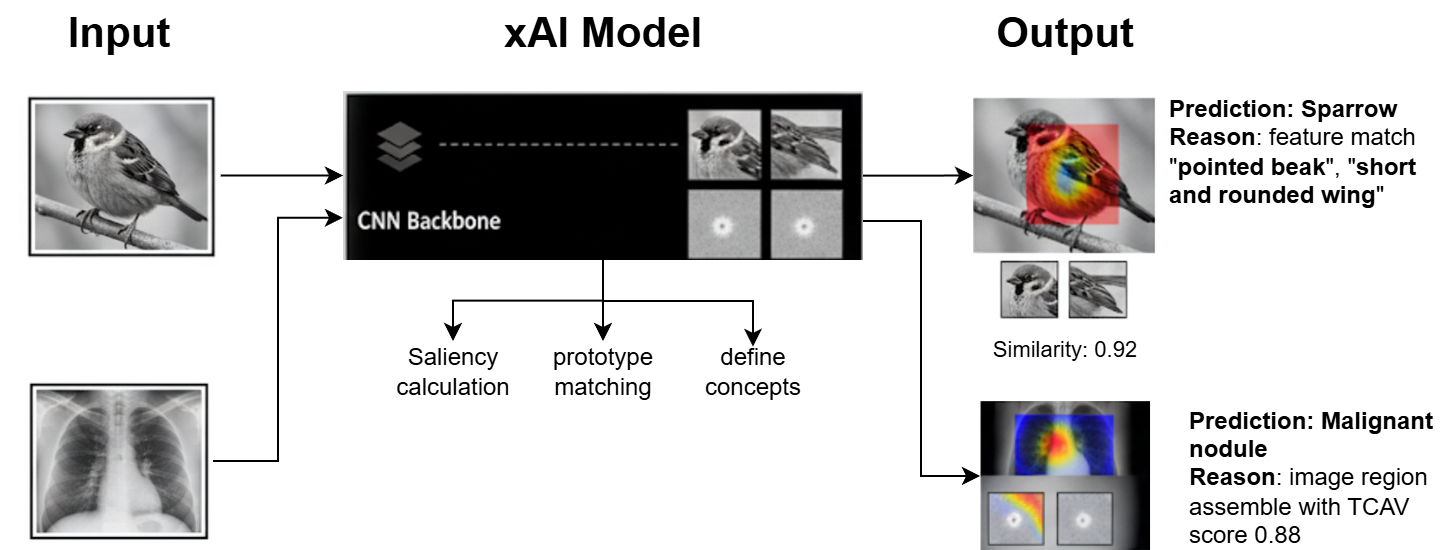}
\caption{Workflow of an explainable AI (xAI) model: input images are processed through a CNN backbone for saliency calculation, prototype matching, and concept definition, leading to predictions with explanations}
\label{fig:xAI_questions}
\end{figure}

As illustrated in Figure \ref{fig:xAI_questions}, the challenge lies not only in building a model capable of making accurate predictions, but also in answering a series of important questions that follow those predictions: \textbf{"Why did the model arrive at this conclusion?"}, \textbf{"How can we trust it?"}, and \textbf{"How can we detect and improve when the model makes mistakes?"}. The shift from a system that only provides “answers” to one that can engage in an explanatory “dialogue” is the foundation of xAI. It helps transform the model from a mere tool into a trusted partner. \\

An ideal xAI system should not only explain why a prediction is correct, but also be able to indicate why the model might be wrong. This is a crucial expectation, especially in real-world applications where errors can have serious consequences. Therefore, a comprehensive explanation must meet the following criteria: (i) indicate the features, evidence, or concepts that the model relied on to make its decision; (ii) reflect the level of confidence and signal when the model is uncertain about its prediction; and (iii) provide useful information so that users, such as doctors, can identify the cause of the error and intervene in a timely manner. \\

However, the path to achieving such an ideal xAI system is fraught with systemic challenges. First, \textbf{explanations must ensure fidelity}, meaning they must accurately reflect the model's internal reasoning mechanisms, rather than merely providing a seemingly plausible but inaccurate justification \cite{zhang2021trust}. Second, \textbf{explanations need to be stable and robust}, as minor changes in the input data should not lead to large and unpredictable changes in the explanation results \cite{brankovic2024benchmarking}. Third, \textbf{the scope of explanations is currently limited}, as most studies focus only on explaining correct predictions while neglecting to analyze the causes of model failure \cite{bove2024fail}. Finally, \textbf{the lack of common metrics and evaluation standards} poses significant challenges in fairly and objectively comparing different methods \cite{ghanvatkar2024evaluating}. \\

To address these challenges and move closer to the ideal goal, the research community has developed a variety of approaches. This survey will systematize and analyze four representative groups of methods, each representing a different explanatory philosophy:
\begin{itemize}
\item \textbf{(i) Saliency Maps.} Answer the question “Where does the model look?” by generating heatmaps that highlight important image regions at the pixel level.
\item \textbf{(ii) Concept Bottleneck Models (CBM).} Answer the question “What is the model thinking?” by forcing the model to reason through high-level concepts that humans can understand.
\item \textbf{(iii) Prototype-based Methods.} Answer the question “What case does this data resemble?” by explaining predictions through comparison with prototypes learned from the dataset.
\item \textbf{(iv) Hybrid Approaches.} Combine the above methods to leverage the strengths of each and provide a more multifaceted, comprehensive explanation.
\end{itemize}

Recognizing the complex context created by the aforementioned challenges, this survey is structured around three main objectives: (1) to systematize and deeply analyze representative xAI methods in computer vision; (2) to synthesize and discuss the evaluation metrics currently in use; and (3) to identify remaining challenges and promising future directions, aiming towards truly reliable xAI systems for real-world applications.

\section{Methodology}
\begin{figure}[H]
\centering
\includegraphics[width=0.8\textwidth]{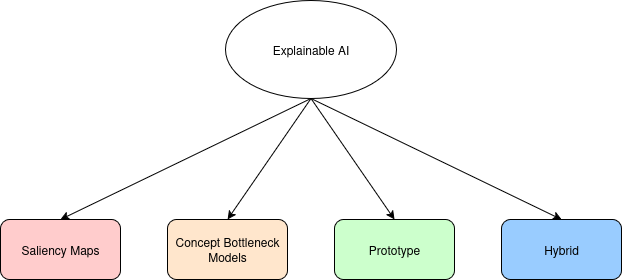}
\caption{Taxonomy of main xAI methods for computer vision}
\label{fig:taxonomy}
\end{figure}

In this section, we present four representative approaches in the field of xAI for computer vision, 
including: \textbf{(i) Saliency Maps}, \textbf{(ii) Concept Bottleneck Models}, \textbf{(iii) Prototype-based Methods}, 
and \textbf{(iv) Hybrid Approaches}. These methods are illustrated in Figure~\ref{fig:taxonomy}, 
showing the taxonomy we constructed to systematize and clarify the role of each group.

\subsection{Saliency Maps}

Saliency maps are a popular group of methods in xAI, used to visualize the regions in an image that the model considers most important for its prediction. These methods are primarily post-hoc, meaning they are applied after the model has been trained, without requiring changes to the architecture or training process. Instead of designing models to be self-explanatory, post-hoc methods analyze internal signals (such as gradients, attention) or external influences (perturbation input) to provide explanations for the predictions of a “black box” model.

\subsubsection{Development stages of the method}

Methods for interpreting computer vision models, also known as saliency maps, originate from the foundational idea of Simonyan et al. \cite{simonyan2013saliency}. This technique uses the gradient of the classification score propagated backward to each pixel of the input image to estimate their “sensitivity”. The strength of this approach lies in its simplicity and lack of requirement for model architecture changes. However, saliency maps generated from raw gradients are often noisy and lack sharpness, making visual interpretation difficult. \\

\begin{figure}[H]
    \centering
    \includegraphics[width=0.75\linewidth]{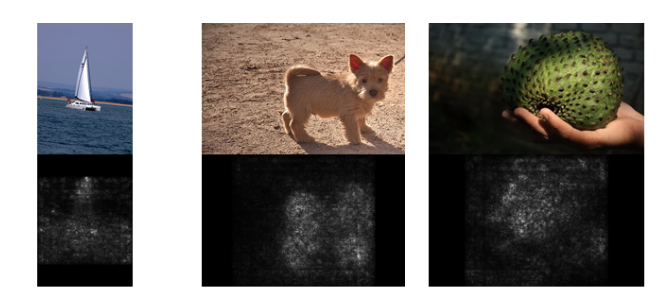}
    \caption{Class-specific saliency maps corresponding to the top-1 predicted category using a single backpropagation pass through a classification ConvNet. The training relied solely on image labels, with no additional annotations employed. \cite{simonyan2013saliency}}
    \label{fig:Saliency Maps}
\end{figure}

To address the issue of noise and improve clarity, a series of improvements based on backpropagation techniques have emerged. Methods such as DeconvNet \cite{zeiler2014deconvnet} and Guided Backpropagation \cite{springenberg2014guidedbp} propose mechanisms that propagate only positive gradients, helping to eliminate noise signals and produce maps that focus more on the activating features of the object. Concurrently, techniques like SmoothGrad \cite{smilkov2017smoothgrad} and Integrated Gradients \cite{sundararajan2017ig} approach the problem from the perspective of stability and consistency. SmoothGrad reduces noise by averaging maps from multiple lightly noisy versions of the original image, while Integrated Gradients ensures fair weight distribution for each pixel by integrating the gradient along a baseline. \\

A true breakthrough came from shifting from pixel-level analysis to semantic feature-level analysis. Zhou et al. \cite{zhou2016cam} introduced Class Activation Maps (CAM), a method for generating heatmaps by taking the weighted sum of feature maps from the final convolutional layer. Although highly effective, CAM requires a specific network architecture with a Global Average Pooling layer, limiting its applicability. This limitation was completely removed by Selvaraju et al. \cite{selvaraju2017gradcam} with Grad-CAM. By using gradients to calculate weights for feature channels, Grad-CAM becomes a general method that can be applied to most CNN architectures without any modifications. This is considered an important milestone, enabling saliency maps to be both semantically meaningful and widely applicable. \\

\begin{figure}[H]
    \centering
    \includegraphics[width=0.75\linewidth]{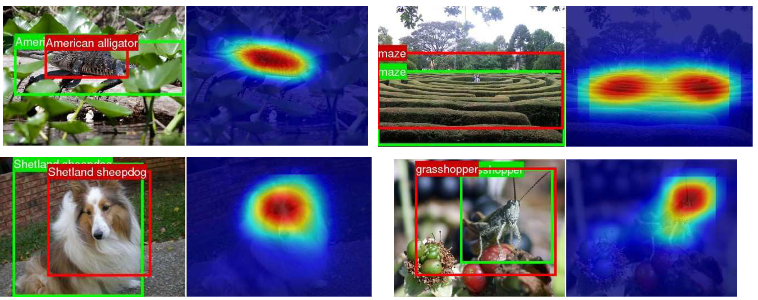}
    \caption{A simple modification of the global average pooling layer combined with class activation mapping (CAM) technique allows the classification-trained CNN to both classify the image and localize class-specific image regions in a single forward-pass \cite{zhou2016cam}}
    \label{fig:Class Activation Maps}
\end{figure}

As Grad-CAM became the standard, subsequent studies focused on addressing its remaining weaknesses. Grad-CAM++ \cite{chattopadhay2018grad} was proposed to improve localization accuracy in complex scenarios, such as when multiple objects of the same class are present in an image. By using a weighted average of positive gradients, Grad-CAM++ generates maps that cover the entire object more accurately. Another inherent limitation is the low resolution of heatmaps, as they are generated from the final feature layer. Layer-CAM \cite{jiang2021layercam} addresses this issue by proposing a technique to generate and combine activation maps from multiple layers at different depths. Combining information from shallow layers (high resolution, detailed) and deep layers (semantically rich) enables Layer-CAM to generate heatmaps that are both sharp and accurately focused on the object. \\

The rise of attention-based architectures, particularly the Vision Transformer (ViT) \cite{dosovitskiy2020image}, has opened a new chapter for interpretation methods. Since ViT lacks spatial convolution layers, CAM-based techniques are no longer suitable. Instead, researchers have leveraged the Transformer's self-attention mechanism itself. The initial approach was to aggregate attention matrices across layers to create an overall map, with the representative technique being Attention Rollout \cite{abnar2020quantifying}. However, Chefer et al. \cite{chefer2021transformer} pointed out that attention alone is insufficient and can be misleading. They developed a more comprehensive method, combining both attention scores and information from gradients to calculate “relevance scores”. This technique produces more accurate and reliable interpretation maps, paving the way for understanding and trusting Transformer models in critical applications.\\

\subsubsection{Practical Applications}
Saliency maps have been widely applied in many biomedical problems, particularly in medical image diagnosis tasks. This method provides a visual tool to help physicians better understand the decision-making mechanism of deep learning models, thereby increasing transparency and reliability in clinical applications. \\

One notable application is \textbf{glaucoma diagnosis via fundus images}. 
In the study \cite{schlemper2023glaucoma}, the authors used multiple saliency methods to identify the important regions that CNN relies on when predicting pathology. The results show that, although saliency maps provide useful visual information, there are still significant differences compared to the annotation standards of ophthalmology experts.  \\

For \textbf{chest X-rays}, many studies focus on evaluating the reliability of saliency maps. The study \cite{arun2020assessing} indicates that common methods like Grad-CAM often fail to meet certain criteria such as reproducibility or sensitivity to weight changes, thereby posing challenges 
regarding reliability in practical applications.  \\

Additionally, a new development direction is the use of \textbf{attention-based saliency maps} in Vision Transformers (ViTs). The study \cite{kim2022pneumothorax} shows that attention-based maps can better reproduce abnormal regions than Grad-CAM in the pneumothorax classification task, while also receiving higher clinical ratings from physicians.  \\

In the context of the COVID-19 pandemic, saliency maps are also used to support \textbf{classifying pneumonia and COVID-19} 
from chest X-rays. The work \cite{zhang2023covid} proposes combining multi-layer Grad-CAM (ML-Grad-CAM) with 
lung ROI processing techniques, achieving high accuracy (96.44\%) while providing visual heatmaps for abnormal regions such as ground-glass opacity (GGO).  \\

In summary, saliency maps have demonstrated practical value in many medical applications. However, studies 
also indicate that their reliability and alignment with medical knowledge remain limited, necessitating future improvements and integration 
with other interpretive methods.

\subsubsection{Measures for Evaluating Saliency Maps}

A key issue in saliency map research is how to evaluate the quality of 
generated maps. Various metrics have been proposed in many studies, which can be divided into 
two groups: quantitative and qualitative. \\

\textbf{Area Over the Perturbation Curve (AOPC).}
Samek et al. \cite{samek2017evaluating} introduced the AOPC metric, based on the idea that if a heatmap accurately reflects 
important regions, then when gradually perturbing the most important pixels, the model's classification score 
will decrease rapidly. AOPC is defined as the average decrease in confidence when perturbing the $k$ most important regions sequentially:

\[
\text{AOPC} = \frac{1}{L} \sum_{k=1}^{L} \left( f(x) - f(x^{(k)}) \right)
\]

Where $f(x)$ is the score of the original image, $f(x^{(k)})$ is the score when $k$ important regions have been corrupted, 
and $L$ is the total number of corruption steps. A high AOPC indicates a useful heatmap.\\

\textbf{Complexity Measures.} Samek et al. \cite{samek2017evaluating} also a metrics for the complexity of the heatmap via entropy.

For a heatmap $H$ normalized as a probability distribution:

\[
\text{Entropy}(H) = - \sum_{i} H_i \log H_i
\]

A good heatmap typically focuses on a few distinct areas, thus having low entropy. Additionally, the file size after compression 
is also used as a secondary indicator. \\

\textbf{Deletion/Insertion.} Fong and Vedaldi \cite{fong2017meaningful} developed a visual metric: in the deletion scenario, 
the most important pixels are gradually removed and the rate of confidence decrease is observed; in insertion, 
important pixels are gradually added to the empty image and the rate of confidence increase is observed. The general formula:

\[
\text{Score} = \frac{1}{L} \sum_{k=1}^{L} f(x^{(k)})
\]

Where $x^{(k)}$ is the image after deleting or inserting the $k$ most important pixels.
A good heatmap will show a rapid decrease in deletion and a rapid increase in insertion. \\

\textbf{Pointing Game.} Zhou et al. \cite{zhou2016cam} and later Petsiuk et al. \cite{petsiuk2018rise} use the Pointing Game metric. 
Accordingly, the pixel with the highest saliency value $\hat{p}$ is compared to the ground-truth bounding box $B$. 
If $\hat{p} \in B$, it is counted as a ``hit’'; otherwise, it is a ``miss’'. The score is:

\[
\text{Accuracy} = \frac{\#\text{hit}}{\#\text{hit} + \#\text{miss}}
\]

Reflects the degree of correspondence between the saliency map and the actual object. \\

\textbf{Sanity Checks.} Adebayo et al. \cite{adebayo2018sanity} do not define a specific formula, 
but instead verify by randomizing model weights or changing data. 
If the saliency map does not change when the model is randomized, then the explanation method is unreliable. \\

\textbf{Qualitative evaluation.} In addition to quantitative measures, many studies still rely on direct observation (qualitative evaluation), 
which involves visually comparing saliency maps with objects in the image. Although subjective, this evaluation method 
provides clear illustrations and is often used alongside quantitative metrics. \\

In summary, the ecosystem of saliency map evaluation metrics is increasingly diverse, ranging from measurements based on 
disturbance, comparison with ground-truth, to reliability verification. These metrics play a 
crucial role in distinguishing which interpretation methods are truly useful and capable of reflecting 
the learning model.

\subsubsection{Advantages and Limitations of Saliency Maps}

The saliency maps method has become fundamental in research on explaining deep learning models 
on image data due to the following advantages:

\begin{itemize}
    \item \textbf{Ease of application.} Most saliency techniques, especially gradient-based ones 
    \cite{simonyan2013saliency}, can be directly implemented on pre-trained CNN models 
    without altering the architecture or training process.
    \item \textbf{Highly intuitive.} Saliency maps display results as heatmaps, 
    which are easily observable by the naked eye, helping users quickly identify the image regions 
    of interest to the model.
    \item \textbf{Diverse improvement methods.} From initial gradient-based techniques 
    to variants such as CAM/Grad-CAM \cite{zhou2016cam,selvaraju2017gradcam}, SmoothGrad 
    \cite{smilkov2017smoothgrad}, or Integrated Gradients \cite{sundararajan2017ig}, 
    saliency maps have seen numerous improvements aimed at enhancing stability and interpretability.
    \item \textbf{Wide-ranging applications.} Saliency maps are used in many fields, 
    from computer vision (object recognition, segmentation) to medicine (highlighting abnormal regions in X-ray or MRI images).
\end{itemize}

However, this method also has some significant limitations:

\begin{itemize}
    \item \textbf{Unstable results.} Saliency maps are often sensitive to 
    input noise or small changes in the model. For example, Adebayo et al. \cite{adebayo2018sanity} 
    showed that many methods still produce similar heatmaps even when model weights are randomized.
    \item \textbf{Difficulty in quantification.} Although many metrics have been proposed (AOPC, Deletion/Insertion, Pointing Game), 
    none is considered an absolute standard, making comparisons between methods 
    challenging.
    \item \textbf{Risk of Misinterpretation.} Heatmaps are often blurry and diffuse, leading to the risk 
    that users may over-interpret highlighted regions \cite{samek2017evaluating}.
    \item \textbf{Lack of conceptual level.} Saliency maps only reflect important pixel regions, 
    but do not clearly indicate what concepts the model has learned (e.g., “bird beak” or just “bright area”).
\end{itemize}

In summary, saliency maps are one of the first and important steps in the field of xAI for image data, 
but to achieve stronger and more reliable interpretability, additional concept-based 
or prototype-based methods are needed to supplement the inherent limitations of this approach.

\subsection{Concept Bottleneck Models}
\textbf{Concept Bottleneck Models (CBM)} is an approach to interpreting deep learning models 
based on inserting an intermediate layer (\textit{bottleneck layer}) consisting of predefined \textit{concepts} 
defined by humans. The main idea was introduced by Koh et al.
\cite{koh2020concept}, instead of the model directly mapping from input data $x$ to label $y$,
the learning process is split into two steps: 

\[
x \;\;\longrightarrow\;\; c \;\;\longrightarrow\;\; y
\]

where $c$ is the vector of intermediate concepts.  
\begin{itemize}
    \item Step 1: The model predicts the values of the concepts $c$ from the input data. 
    \item Step 2: These concepts $c$ are used to predict the final label $y$. 
\end{itemize}
  
This approach makes interpretation more transparent, as users can directly observe 
the contribution level of each concept in the model's reasoning process. For example, instead of simply indicating 
“dimly lit area”, CBM can explain that the prediction stems from the concept \textit{opacity} or 
\textit{cardiomegaly}.  

\subsubsection{Development stages of the method}

A breakthrough in building interpretable-by-design models came from the idea of Concept Bottleneck Models (CBM), first introduced by Koh et al. \cite{koh2020concept}. Instead of learning a direct mapping from input to output, the CBM architecture inserts an intermediate layer (bottleneck) consisting of a set of meaningful, human-defined \textit{concepts}. The learning process is split into two stages: first, the model learns to predict the presence of these concepts from the input data; then, a second, simpler model uses these concept values to make the final prediction. This design offers two core benefits, the ability to interpret the decision-making process transparently and the ability to allow human intervention to correct errors and debug the model.

\begin{figure}[H]
    \centering
    \includegraphics[width=0.5\linewidth]{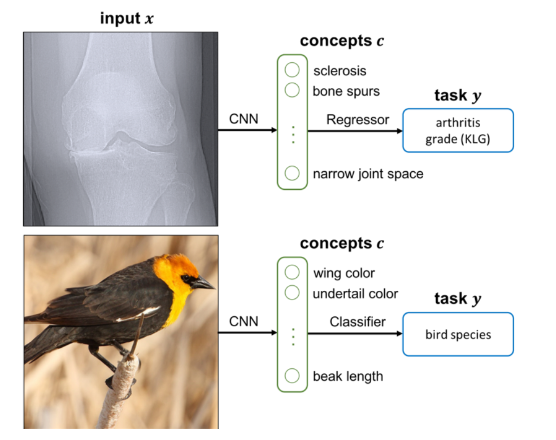}
    \caption{The CBM architecture separates the learning process into two stages: from image to concept and from concept to label \cite{koh2020concept}}
    \label{fig:Class Bottleneck Models}
\end{figure}

However, this groundbreaking idea also comes with significant challenges. A core question is whether the model truly learns and uses concepts in the way humans intend. A study by Margeloiu et al. \cite{margeloiu2021cbmintended} showed that CBMs are not always “faithful”; they can find shortcuts in the data and ignore defined concepts. Furthermore, original CBM faces three major limitations: (1) the extremely costly requirement of manually labeling concepts across large datasets; (2) a trade-off between interpretability and accuracy, as forcing information through a bottleneck can reduce performance compared to end-to-end models; and (3) concepts are only global, lacking detailed spatial information. \\

To address these challenges, the research community has developed several advanced approaches. The first approach focuses on eliminating dependence on concept labels. Methods such as Label-Free CBM \cite{oikarinen2023label} leverage the power of large Vision-Language Models (VLM) like CLIP. Instead of requiring pre-assigned labels, users only need to provide a list of concepts in text form. The model then automatically calculates the similarity between images and these text concepts to create a bottleneck layer, significantly reducing data costs.\\

To address the trade-off in accuracy, Post-hoc CBM (PCBM) \cite{yuksekgonul2022posthoc} has been proposed. Instead of training a CBM model from scratch, PCBM operates on features extracted from pre-trained robust black-box models. In this way, PCBM retains the high performance of the original model while providing concept-based interpretability and intervention capabilities. \\ 

Finally, to enhance the level of detail in explanations, studies have focused on concept localization and quantifying uncertainty. Probabilistic CBM \cite{marconato2023probabilistic} models each concept as a probability distribution, enabling the model to express the degree of uncertainty in its predictions, a critically important factor in high-risk applications. These developments are gradually transforming CBM from a theoretical idea into a more powerful, reliable, and practical tool.

\subsubsection{Practical Applications}
Concept Bottleneck Models (CBMs) have been applied in many biomedical fields, where interpretability plays a 
crucial role in supporting healthcare professionals in decision-making. Unlike saliency maps, which only provide visual information 
at the pixel level, CBMs are based on semantic concepts, thereby providing explanations that are closer 
to human reasoning. \\  

A notable application is in \textbf{dermatological image diagnosis}. The study \cite{lampert2020concept} built 
a CBM for classifying skin lesions, where concepts such as \textit{“red color”, “abnormal border”} 
were learned as intermediate layers before classifying malignancy or benignity. This enables physicians to observe 
the model based on clinical concepts to make diagnoses.  \\

In \textbf{chest X-rays}, \cite{yuksekgonul2022posthoc} proposed a method using post-hoc CBM 
to examine the association between medical concepts (e.g., \textit{cardiomegaly, opacity, effusion}) 
and disease prediction. Results show that CBMs not only achieve high predictive performance but also provide 
intervenability physicians can directly modify concept values (e.g., reassign the state of “effusion”) 
to observe changes in classification outcomes. \\ 

Additionally, in the \textbf{multimodal explanation task}, some studies combine CBM with natural language 
to provide textual explanations for medical images. For example, \cite{yuksekgonul2023posthoc} combines CBM with LLM 
to generate concept-based explanatory descriptions, enhancing user-friendliness for non-specialists.  \\

In summary, CBMs have demonstrated potential in bridging the gap between deep learning models and the reasoning 
processes of medical experts, particularly through their semantic concept-based capabilities and support for direct intervention. However, 
a major challenge remains in accurately identifying and labeling medical concepts, which requires significant costs 
and the involvement of medical specialists.

\subsubsection{Evaluation Metrics for Concept Bottleneck Models}

The evaluation of Concept Bottleneck Models (CBM) focuses on two main aspects: 
(1) accuracy in predicting concepts and final labels, 
and (2) the extent to which the model actually learns and uses the desired concepts. 
During development, numerous metrics have been proposed, ranging from basic indices to 
more complex and comprehensive metrics. \\

\textbf{Original CBM (Koh et al., 2020).} In the original work \cite{koh2020concept}, two main metrics were used:
\begin{itemize}
    \item \textbf{Task Accuracy.} The accuracy of the final output label $y$, calculated using the error rate (0-1 error) or RMSE:
    \[
    RMSE(y, \hat{y}) = \sqrt{\frac{1}{n}\sum_{i=1}^n (y_i - \hat{y}_i)^2}
    \]
    \item \textbf{Concept Accuracy.} The accuracy when predicting concepts $c$, also calculated using RMSE or 0-1 error.
\end{itemize}

\textbf{Evaluating learning behavior (Margeloiu et al., 2021).} The study \cite{margeloiu2021cbmintended} asks whether CBM actually learns the correct concept. 
The metrics used include:
\begin{itemize}
    \item \textbf{Intervenability Test.} Replace the predicted concept with the ground-truth concept and measure the improvement in 
    accuracy. If the result increases significantly, it proves that the model truly relies on the concept.
    \item \textbf{Saliency Map Alignment (Qualitative).} Visually compare the concept heatmap with the ground-truth region.
    \item \textbf{Quantitative Check.} Measure the correlation coefficient $R^2$ between the saliency map from CBM and from Concept Oracle:
    \[
    R^2 = 1 - \frac{\sum_i (h_i - \hat{h}_i)^2}{\sum_i (h_i - \bar{h})^2}
    \]
\end{itemize}

\textbf{Standardization and Extension (Aysel et al., 2025).} The recent work \cite{aysel2025conceptmetrics} proposes a more standardized set of metrics, including CGIM, CEM, and CLM. \\

\textbf{Concept Global Importance Metric (CGIM).} Evaluates the similarity between the concept importance learned by the model 
and that defined by humans, using cosine similarity:

\[
\rho_{CGIM} = \cos(\theta_j, V_j) = \frac{\theta_j \cdot V_j}{\|\theta_j\| \|V_j\|}
\]

where $\theta_j$ is the concept weight vector learned by the classifier for concept $j$, 
and $V_j$ is the ground-truth concept vector provided by humans. CGIM has three main variants:
\begin{itemize}
    \item \textbf{CGIM1.} directly compares the weight vector $\theta(j,:)$ with $V(j,:)$.
    \item \textbf{CGIM2.} compares the average predicted concept value $\hat{U}^*(j,:)$ with $V(j,:)$.
    \item \textbf{CGIM3.} Combines both the weights and the average predicted value 
    by element-wise multiplication $\hat{U}^* \odot \theta(j,:)$ and then compares it with $V(j,:)$.
\end{itemize}

The value $\rho_{CGIM}$ lies in the range $[-1,1]$. 
If $\rho_{CGIM}$ is close to $1$, it indicates that the model and humans agree strongly 
on which concepts are important for that class. 
If the value is low or negative, the model may be relying on concepts 
that do not accurately reflect the essence that humans expect. \\

\textbf{Concept Existence Metric (CEM).} CEM is designed to check whether the most important concepts that the model selects for an image 
actually exist in that image according to the ground truth. 
The symbol $\Lambda_i$ is the set containing the true concept indices (ground truth) 
of image $X_i$. With $q_i$ being the model's concept importance vector and $l \leq L$ 
being the number of most important concepts considered, CEM is defined as follows:

\[
\rho^{CEM}_l := \frac{1}{l} \sum_{j=1}^{l} \mathbf{1}_{\Lambda_i}(q_{ij}),
\]

where the indicator function $\mathbf{1}_{\Lambda_i}(x)$ is defined as follows:

\[
\mathbf{1}_{\Lambda_i}(x) = 
\begin{cases}
1, & \text{if } x \in \Lambda_i \\
0, & \text{otherwise.}
\end{cases}
\]
\\
\textbf{Concept Location Metric (CLM).} CLM is designed to evaluate the location of a concept in an image. 
Specifically, it checks whether the concept-wise heatmap $F_{ij}$ for concept $j$ 
generated by CoAM contains the ground-truth center position $p_{ij}$. 
For the $l \leq L$ most important concepts of $X_i$, CLM is defined as follows:

\[
\rho^{CLM}_l := \frac{1}{l} \sum_{j=1}^{l} \mathbf{1}_{\Omega_{ij}}(p_{ij}),
\]

where $\mathbf{1}_{\Omega_{ij}}(p_{ij}) = 1$ if the position $p_{ij}$ belongs to the activation region $\Omega_{ij}$ 
of concept $j$, and equals $0$ otherwise. \\

In summary, the metrics system for CBM has evolved from basic indicators such as accuracy, 
to intervention validation and visualization, and most recently to a set of quantitative standards 
with CGIM, CEM, and CLM for more comprehensive evaluation.

\subsubsection{Advantages and Limitations}
Concept Bottleneck Models (CBMs) offer a more structured and transparent approach, 
addressing many of the conceptual shortcomings found in methods like Saliency Maps. The main advantages of this approach include:

\begin{itemize}
    \item \textbf{Interpretability through concepts.} CBMs provide human-understandable explanations 
    by reasoning in terms of high-level concepts rather than abstract features, making the model’s 
    decision-making process more transparent.
    \item \textbf{Intervention and error correction capability.} A key feature of CBMs is allowing users 
    to intervene directly at the concept layer. This enables error correction, hypothesis testing, 
    and strengthens trust in the model for critical applications, something that many post-hoc methods cannot offer.
    \item \textbf{Potential for causal reasoning.} By explicitly modeling the reasoning chain from 
    input to concepts to output, CBMs open up the potential to move beyond correlation-based explanations 
    toward more robust causal and counterfactual analyses.
\end{itemize}

However, despite their innovative approach, CBMs also present several significant theoretical and 
practical challenges:

\begin{itemize}
    \item \textbf{Burden of concept definition.} The effectiveness of a CBM heavily depends on a 
    predefined set of concepts. This process requires significant human effort, deep domain expertise, 
    and is inherently subjective in identifying truly meaningful concepts.
    \item \textbf{Risk of shortcut learning.} The model may learn to exploit spurious correlations 
    between concepts and the final labels. This can lead to explanations that appear correct but 
    are not faithful to the human-intended reasoning process.
    \item \textbf{Scalability challenges.} In domains with a large number of abstract or ambiguous concepts, 
    the process of defining, annotating, and training for each concept becomes impractical, 
    limiting the method's scalability.
\end{itemize}

In summary, Concept Bottleneck Models represent a significant step toward a more transparent and 
interactive AI by shifting the focus from low-level features to high-level, human-understandable concepts. 
However, the reliance on manual concept engineering remains a major barrier to widespread adoption. 
Future research could focus on automating the concept discovery process or developing hybrid models 
to better balance interpretability and practicality.

\subsection{Prototype-based Methods}

Prototype-based methods are an ante-hoc\footnote{Methods that are designed to be interpretable before model training or deployment, as opposed to post-hoc which explain models after training.} approach to model interpretation by comparing input data with a set of “prototypes” that the model has learned from the training set. Instead of explaining based on gradients or abstract concepts, this approach reasons along the lines of “this object looks like that object”, which is very close to human thinking.

\subsubsection{Development stages of the method}

\begin{figure}[H]
\centering
\includegraphics[width=0.85\linewidth]{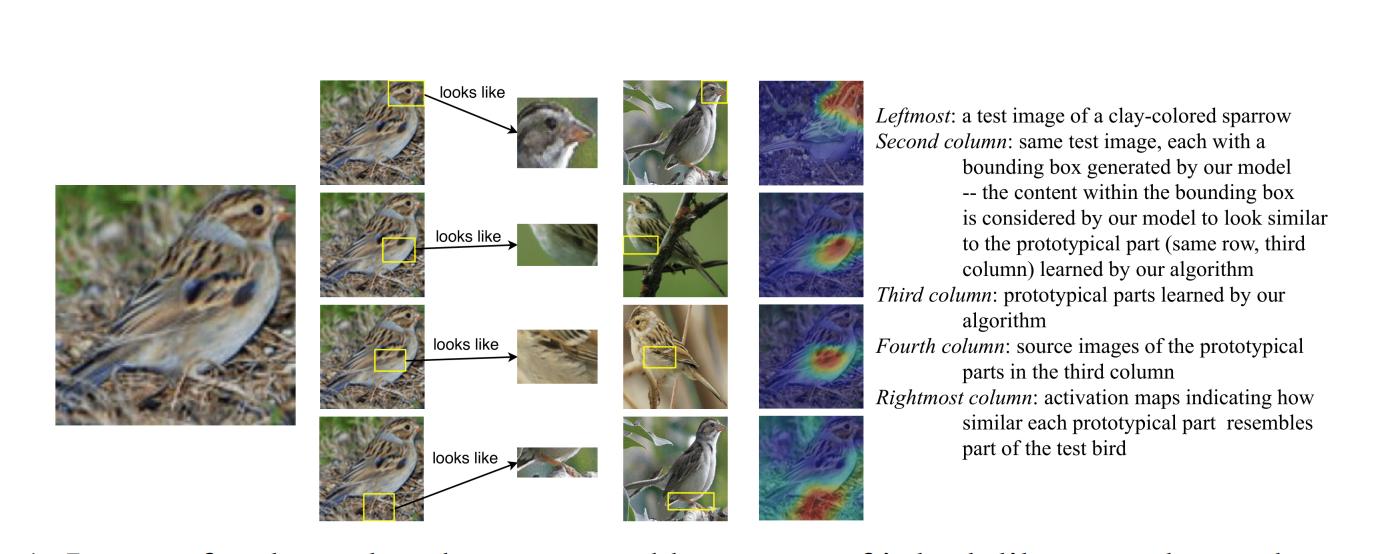}
\caption{ProtoPNet explain mechanisim \cite{chen2019looks}}
\label{fig:protopnet}
\end{figure}

The foundational work for this approach is \textbf{ProtoPNet} by Chen et al. \cite{chen2019looks}. The architecture of ProtoPNet consists of a standard CNN, followed by a prototype layer and finally a linear layer. During training, the model learns a set of prototypes, where each prototype is an image patch sample representing a feature of a specific class. When predicting on a new image, the model will:
\begin{enumerate}
\item Calculate the similarity between regions in the feature map of the input image and each learned prototype.
\item Identify the strongest activated image regions for each prototype.
\item The final linear layer will make a prediction based on the similarity scores of the prototypes.
\end{enumerate}
The explanation results include indicating which regions on the input image resemble which prototype, along with an image of that prototype. For example in Figure \ref{fig:protopnet}, the model might conclude “this is a sparrow” because “its head resembles this sparrow head prototype”.

However, ProtoPNet has some limitations, such as prototypes that may not be “pure” (i.e., representing multiple concepts or containing background noise). To overcome this, subsequent studies have proposed several improvements:
\begin{itemize}
    \item \textbf{ProtoTree.} \cite{prototree2021} Combines prototypes with decision trees to generate a more structured and logical explanation chain (“this image has a pointed beak AND a crested head, therefore it is bird A”).
    \item \textbf{ProtoPool.} \cite{protopool2022} Instead of rigidly assigning each class to a fixed number of prototypes, ProtoPool proposes a new “prototype pool” mechanism for each class and uses a soft weight assignment mechanism (differentiable assignment). This allows the model to select the most suitable combination of prototypes from the “pool” to explain a specific image, enhancing the diversity and flexibility of the explanation.
    \item \textbf{PIP-Net.} \cite{pipnet2023} This approach focuses on learning more “intuitive” prototypes by forcing the model to make decisions based on the presence of specific “patches” (image regions). Instead of learning abstract prototype vectors, PIP-Net learns to score patches in an image, and the patches with the highest scores from the training set become prototypes. The explanation becomes very direct: “this image is classified as X because it contains patches A, B, and C”.
\end{itemize}

\subsubsection{Applications}
Thanks to their transparency and intuitiveness, prototype-based methods have many valuable applications, particularly in fields requiring high reliability and interpretability: \\

\textbf{Medicine and Medical Imaging.} This is one of the most important application areas. Models like ProtoPNet can be used to classify medical images (e.g., X-rays, histopathology images) and explain diagnostic reasons. For example, the model might conclude that a tumor is malignant because “its cellular structure resembles this cancer cell prototype”. This not only helps doctors trust the results but can also help them detect morphological features that humans might overlook. \\

\textbf{Biological and environmental classification. }In extremely detailed animal or plant species classification tasks (fine-grained recognition), these methods help biologists understand which morphological features (e.g., beak shape, feather color) the model relies on to distinguish between similar bird species. \\

\textbf{Self-driving cars and robots. }In self-driving cars, understanding why an object is identified as a “pedestrian” or a “stop sign” is critically important for safety. Prototype-based models can explain that “this object was identified as a pedestrian because its torso and leg shape resemble the prototype of a walking person”. This aids in debugging the system and ensures the model does not rely on misleading signals. \\

\textbf{Quality control in manufacturing. }Models can be trained to detect defective products on the production line. The explanation based on the prototype will specify “this product is defective because it has a scratch similar to the scratch defect prototype”.

\subsubsection{Evaluation metrics}
Evaluating prototype-based methods typically revolves around the following aspects, with specific mathematical formulas: \\

\textbf{Task Accuracy.} This is the most fundamental metric, evaluating the model's performance on the primary task (e.g., classification).
\[
\text{Accuracy} = \frac{\sum_{i=1}^{N} \mathbf{1}(\hat{y}_i = y_i)}{N}
\]
In which:
\begin{itemize}
    \item $N$ is the total number of samples in the evaluation set.
    \item $y_i$ is the true label of the $i$th sample.
    \item $\hat{y}_i$ is the predicted label of the $i$th sample.
    \item $\mathbf{1}(\cdot)$ is the indicator function, returning 1 if the condition inside is true, and 0 otherwise.
\end{itemize}

\textbf{Prototype Purity.} This measure evaluates whether each prototype truly represents a single concept/class. One quantitative approach is to calculate the proportion of the class that dominates the strongest activating samples in a prototype.
\[
\text{Purity}(p_k) = \max_{j \in \{1, ..., C\}} \frac{|\{x_i \in D_k \mid y_i = j\}|}{|D_k|}
\]
Where:
\begin{itemize}
    \item $p_k$ is the $k$th prototype.
    \item $D_k$ is the set of training samples $x_i$ for which $p_k$ is the most strongly activated (closest) prototype.
    \item $C$ is the total number of classes.
    \item $y_i = j$ indicates that sample $x_i$ belongs to class $j$.
    \item $|\cdot|$ denotes the number of elements in the set.
\end{itemize}
The overall purity of the model is the average value of $\text{Purity}(p_k)$ across all prototypes $k$. \\

\textbf{Fidelity.} Check whether the explanation truly reflects the model's decision-making process, often measured using perturbation techniques. For example, we can use the AOPC (Area Over the Perturbation Curve) index by occluding the image region corresponding to the most important prototype and observing the decrease in the prediction score.
\[
\text{AOPC} = \frac{1}{L} \sum_{k=1}^{L} \left( f(x) - f(x^{(k)}) \right)
\]
In which:
\begin{itemize}
    \item $f(x)$ is the predicted score of the model for the correct class on the original image $x$.
    \item $x^{(k)}$ is the image after masking the region corresponding to the $k$ prototypes with the greatest influence.
    \item $L$ is the total number of perturbation steps.
\end{itemize}
A high AOPC value indicates that the selected prototypes are truly important for the model's decision. \\

\textbf{Localization Accuracy.} If the problem has available bounding boxes for objects or parts, we can evaluate the degree of overlap between the image region that activates the prototype and that bounding box.

\begin{itemize}
    \item \textbf{Pointing Game.} Evaluate whether the point with the highest activation value lies within the bounding box.
    \[
    \text{Accuracy}_{\text{PG}} = \frac{1}{N} \sum_{i=1}^{N} \mathbf{1}(\hat{p}_i \in B_i)
    \]
    Where $\hat{p}_i$ is the pixel coordinate with the highest prototype activation value in image $i$, and $B_i$ is the ground-truth bounding box.
    
    \item \textbf{Intersection over Union (IoU).} Measures the overlap between the activation region and the bounding box.
    \[
    \text{IoU}(A, B) = \frac{|A \cap B|}{|A \cup B|}
    \]
    Where $A$ is the image region identified by the model as important (e.g., activation region above a certain threshold) and $B$ is the ground-truth bounding box.
\end{itemize}
  
\subsubsection{Advantages and Limitations}
The prototype-based approach offers an intuitive way of explaining model decisions by comparing a new input 
to representative examples (prototypes) learned from the training data. This approach mimics human-like reasoning 
and presents several notable advantages:

\begin{itemize}
    \item \textbf{High intuitiveness.} Explanations using concrete examples are highly intuitive because 
    they closely resemble human reasoning, comparing an unknown object to familiar ones.
    \item \textbf{Ante-hoc by design.} The explanation mechanism is built directly into the model's architecture, 
    ensuring that the provided reasons faithfully reflect the model's actual reasoning process, rather than 
    being a post-hoc interpretation.
    \item \textbf{Provides real-world examples.} Users can directly examine the prototypes, which are actual 
    samples from the training data. This provides grounding and clear context for the model's decisions.
\end{itemize}

However, building an explanation mechanism based on prototypes also comes with inherent limitations:

\begin{itemize}
    \item \textbf{Requires architectural modification.} This method cannot be applied directly to pre-trained 
    models. It requires designing or modifying the architecture from the outset to incorporate the prototype layer.
    \item \textbf{Difficulty in learning meaningful prototypes.} During training, prototypes are at risk of 
    learning noisy features or spurious correlations from the data, resulting in them not representing a 
    truly clear or distinct concept.
    \item \textbf{High computational cost.} The inference process requires comparing the input against all 
    prototypes, which can be computationally expensive, especially when a large number of prototypes is needed 
    to cover the data's diversity.
\end{itemize}

In summary, the prototype-based approach is a powerful direction for creating faithful and understandable explanations. 
However, the barriers of architectural modification and computational cost make it less flexible than post-hoc techniques. 
Future research might focus on optimizing the comparison process or combining it with other methods to mitigate these drawbacks.
\subsection{Hybrid Approaches}
Hybrid methods were developed to combine the strengths of different approaches, thereby creating more comprehensive, reliable, and flexible explanations. Instead of focusing solely on “which regions are important” (saliency maps), “which concepts are used” (CBM), or “which examples are similar” (prototypes), hybrid methods can answer multiple questions simultaneously.

\subsubsection{Mechanisms and Development Directions}

\textbf{TCAV (Testing with Concept Activation Vectors).} This is one of the most influential post-hoc hybrid methods, introduced by Kim et al. \cite{kim2018tcav}. TCAV combines concept-based and gradient-based methods to quantify the importance of a user-defined concept for a specific prediction class. The process involves the following steps:
\begin{enumerate}
    \item \textbf{Concept definition.} The user provides a set of sample images for a concept (e.g., images with “stripes”) and a set of random images that do not contain that concept.
    \item \textbf{Extracting activation vectors.} These images are passed through a pre-trained model, and the activation vectors at an intermediate layer are extracted.
    \item \textbf{Learning Concept Activation Vectors (CAV).} A linear classifier (e.g., SVM) is trained to distinguish the activation vectors of the concept group and the random group. The normal vector of this separating hyperplane is called the CAV, denoted as $\vec{v}_c$. This vector represents the “direction” of the concept in the model's activation space.
    \item \textbf{Calculating sensitivity using directional derivatives.} For an input image $x$, TCAV calculates the directional derivative of the predicted logit along the direction of the CAV. This indicates whether moving in the activation space along the direction of the “stripes” concept increases or decreases the predicted probability of the “zebra” class.
\end{enumerate}
The final result is a quantitative score indicating the percentage of images in a class that are positively influenced by that concept. \\

\textbf{Combining Ante-hoc and Post-hoc.} This is a common hybrid strategy for validation and reliability enhancement. A model with an existing explanatory mechanism (ante-hoc) such as CBM or ProtoPNet will be further tested using a post-hoc tool.
\begin{itemize}
    \item \textbf{Example 1 (CBM + Saliency Map).} After training a CBM model, we can apply Grad-CAM to a specific concept neuron (e.g., the “bird beak” concept). If the resulting heatmap correctly focuses on the beak region of the bird in the image, this reinforces confidence that the model has learned the concept humans intended.
    \item \textbf{Example 2 (ProtoTree + LIME).} A prediction from ProtoTree can be further explained using LIME to see which local superpixels support that decision. If LIME also highlights the region corresponding to the prototype ProtoTree used, the explanation becomes more reliable.
\end{itemize}

\textbf{Other approaches.} Some studies combine prototypes and concepts, such as CSR (Concept-based Similarity Reasoning). Instead of comparing similarity at the pixel level like ProtoPNet, these methods can compare similarity at the concept level. An explanation might be: “This image is classified as X because its concept vector (e.g., \{beak: 0.9, wing: 0.8, webbed foot: 0.95\}) is very similar to the concept vector of a typical duck”.

\subsubsection{Applications}
The combination of multiple interpretation techniques enables hybrid methods to address complex problems requiring cross-validation and a multi-dimensional perspective: \\

\textbf{Debugging and checking model bias.} TCAV is a powerful tool for checking whether a model relies on unwanted concepts. For example, developers can check whether a job application classification model is negatively affected by the concept of “female gender.” By quantifying the importance of sensitive concepts, organizations can build fairer and more responsible AI systems. \\

\textbf{Human-AI Collaboration.} In specialized fields like healthcare or law, hybrid methods provide a rich set of interpretability tools. A system can simultaneously highlight important image regions (saliency maps), identify similar prototypes, and quantify the importance of medical concepts. This allows experts (doctors, lawyers) to interact with, validate, and even refine the model's reasoning, creating an effective collaborative loop. \\

\textbf{Compliance and Auditing.} In industries such as finance and insurance, legal regulations (e.g., Europe's GDPR) require automated decisions to be explainable. Hybrid methods enable the creation of multi-layered explanatory reports that meet stringent auditing requirements by providing both local and global evidence for a decision. \\

\textbf{Scientific Research and Exploration.} Scientists can use hybrid tools to explore how AI models “think” about complex data. For example, a climatologist could use TCAV to check whether a weather forecasting model has learned meteorological concepts such as “convective flow,” thereby gaining new insights into both the model and the data.

\subsubsection{Evaluation Metrics}
Evaluating hybrid methods is complex because they provide diverse types of explanations. Therefore, metrics often depend on the nature of the hybrid method. \\

\textbf{TCAV Score.} For TCAV, the method itself proposes a quantitative metric. The TCAV Score for a concept $c$ and a class $k$ is defined as the proportion of samples in class $k$ that have positive sensitivity to concept $c$.
\[
\text{TCAV}_{c,k} = \frac{|\{x \in X_k : S_{c,k}(x) > 0\}|}{|X_k|}
\]
Where:
\begin{itemize}
    \item $X_k$ is the set of all samples belonging to class $k$.
    \item $S_{c,k}(x)$ is the sensitivity (conceptual sensitivity) of the prediction for class $k$ to concept $c$ on image $x$, calculated using the directional derivative.
    \item $|\cdot|$ denotes the number of elements in the set.
\end{itemize}
A high score (e.g., 0.9) indicates that concept $c$ is highly important and consistent for classifying class $k$. \\

\textbf{Qualitative and Human Evaluation.} Due to the complexity of the explanation, human evaluation plays an extremely important role. Studies often design user studies to measure:
\begin{itemize}
    \item \textbf{Usefulness.} Does the explanation help experts (e.g., doctors) make more accurate or faster decisions?
    \item \textbf{Understandability.} Do users understand the message conveyed by the explanation?
    \item \textbf{Trustworthiness.} Does the explanation increase users' trust in the model?
\end{itemize}
These results are typically collected through surveys and interviews.

\subsubsection{Advantages and Limitations}

Hybrid approaches were developed to overcome the inherent weaknesses of individual methods (such as Saliency Maps, CBMs, or Prototypes) 
by combining their respective strengths. This direction creates more multifaceted and reliable explanations, 
with several notable advantages:

\begin{itemize}
    \item \textbf{Comprehensive perspective.} Hybrid methods can provide multiple explanatory perspectives simultaneously 
    (e.g., highlighting important image regions while also indicating which concepts are used), 
    enabling users to gain a more holistic and deeper understanding of the model's decision.
    \item \textbf{Improved reliability.} By combining multiple techniques, they allow one method to verify 
    and validate the results of another, thereby increasing the overall trustworthiness of the explanation system.
    \item \textbf{High flexibility.} Many hybrid methods (such as TCAV) are post-hoc in nature, 
    making them applicable to a wide range of pre-trained models without requiring architectural modifications.
\end{itemize}

However, this combination also introduces new technical and interpretative challenges:

\begin{itemize}
    \item \textbf{High complexity.} Implementing, computing, and interpreting results from hybrid methods 
    is often significantly more complex compared to using individual methods.
    \item \textbf{Difficulty in standardizing evaluation.} Due to the diversity of explanations generated, 
    constructing standardized and fair evaluation metrics for hybrid methods remains a major challenge.
    \item \textbf{Requires user effort.} Methods such as TCAV require users to manually define concepts and 
    collect representative samples, which can be a time-consuming and subjective process.
\end{itemize}

In summary, hybrid approaches represent a natural and crucial evolution in the field of xAI, aiming for more 
comprehensive and robust explanation systems. However, the trade-off between the depth of explanation and 
implementation complexity is a significant barrier. The future of xAI may lie in developing standardized 
hybrid frameworks that simplify integration and evaluation, thereby bringing AI explanations closer to end-users.

\section{Discussion}
This survey has systematized and analyzed four main groups of methods in the field of xAI for computer vision. Through this, we can see a comprehensive picture of the field’s development, ranging from visualizing pixel-level influence regions to building inherently interpretable models based on concepts and examples. In this section, we discuss two main aspects: the contributions and core capabilities that these approaches have provided, and the open systemic challenges that the entire field is facing. \\

The groups of xAI methods, each with their own philosophical foundations, have provided a multifaceted toolkit for interpreting model behavior, with each method answering a unique interpretability question. At the most basic level, Saliency Maps address the question "Where does the model look?". Methods such as Grad-CAM \cite{selvaraju2017gradcam} have become the most widely adopted tool for identifying image regions most influential to predictions, serving as a foundational diagnostic step for debugging and detecting spurious features. Going a step further, Concept Bottleneck Models (CBM) answer the question "What concepts does the model reason with?". The key contribution of CBM is the introduction of intervention capability \cite{koh2020concept}, which allows humans to interact, correct errors, and test causal hypotheses inside the model—an essential factor for building trust in specialized domains. In contrast to such abstract explanations, Prototype-based Methods answer the question "Which example does this instance most resemble?". By explaining through prototypical examples, these methods help reveal the data distributions the model has learned and identify boundary cases \cite{chen2019looks}. Finally, recognizing the blind spots of individual methods, Hybrid Approaches emerged to provide more comprehensive explanations, combining the spatial localization strength of Saliency Maps with the semantic reasoning of CBM to simultaneously answer both the "where" and "why" questions. \\

Despite these significant advances, xAI still faces systemic challenges. The foremost challenge is the \textbf{fidelity and robustness} of explanations. A major open question is whether explanations truly reflect the model’s internal reasoning process. Saliency maps, for instance, have been shown to be unstable, as small changes in the input image can lead to large changes in the heatmap \cite{brankovic2024benchmarking}. Even CBMs can learn “shortcuts” and fail to faithfully represent human-defined concepts \cite{margeloiu2021cbmintended}. Most current xAI methods excel at highlighting correlations, but not at uncovering causal relationships. \\

Another challenge is the \textbf{problem of evaluation and standardization}. Quantifying whether an explanation is “good” remains difficult due to the lack of standardized metrics and benchmark datasets \cite{ghanvatkar2024evaluating}. This makes it hard to fairly compare methods. Furthermore, \textbf{human dependence and scalability} remain practical barriers. Powerful approaches like CBM and prototype-based methods often require significant human involvement in defining concepts or interpreting prototypes, reducing scalability. \\

Addressing these systemic challenges points toward several key directions for future research. \textbf{First, the emphasis will shift from correlation-based explanations toward causal reasoning.} Rather than merely identifying relevant features, future methods need the ability to generate counterfactual explanations that can pinpoint the true causal factors driving model decisions. \textbf{Second, the field must urgently establish comprehensive and standardized evaluation frameworks.} Such benchmarks should not only include computational measures of fidelity and robustness but also incorporate human-computer interaction studies to assess the usefulness and effectiveness of explanations for end-users. Finally, \textbf{to enhance scalability and reduce reliance on domain experts, the inevitable direction is the development of methods capable of automatically discovering meaningful concepts and prototypes}, while also constructing principled hybrid models that harmonize spatial accuracy with semantic richness. Addressing these challenges will be key to transforming xAI from a laboratory diagnostic tool into an indispensable partner in real-world critical AI systems.

\section{Conclusion}
This survey has provided a systematic overview of four major families of methods in xAI for computer vision: Saliency Maps, Concept Bottleneck Models, Prototype-based Methods, and Hybrid Approaches. Our primary contribution is the structured analysis and synthesis of these methods, clarifying their underlying philosophies, strengths, limitations, and practical applications. Through this, the paper highlights the field's evolution from answering the basic question of "Where does the model look?" to addressing more complex queries like "What concepts does the model reason with?". Our analysis emphasizes that no single method is a panacea; the choice of an appropriate technique depends on the specific interpretability requirements of the application. Fundamentally, this paper offers a framework to help researchers and practitioners navigate the current xAI landscape and make informed decisions. In closing, the trajectory of xAI is clearly moving toward building systems that are not just transparent but genuinely trustworthy, enabling the safe and responsible deployment of AI in critical domains of society.

\bibliographystyle{unsrt} 
\bibliography{references}

\begin{thebibliography}{10}

\bibitem{zhang2021trust}
Y~Zhang, H~Chao, MK~Kalra, and P~Yan.
\newblock Overlooked trustworthiness of explainability in medical ai.
\newblock {\em medRxiv}, 2021.

\bibitem{brankovic2024benchmarking}
L~Brankovic and Others.
\newblock Benchmarking the most popular xai used for explaining clinical predictive models: Untrustworthy but could be useful.
\newblock {\em Journal of Biomedical Informatics}, 2024.

\bibitem{bove2024fail}
A~Bove, T~Laugel, MJ~Lesot, and M~Detyniecki.
\newblock Why do explanations fail? a typology and discussion on failures in xai.
\newblock {\em arXiv preprint arXiv:2405.13474}, 2024.

\bibitem{ghanvatkar2024evaluating}
S~Ghanvatkar, B~Bensaude~Vincent, and Others.
\newblock Evaluating explanations from ai algorithms for clinical decision-making: A social science-based approach.
\newblock {\em medRxiv}, 2024.

\bibitem{simonyan2013saliency}
Karen Simonyan, Andrea Vedaldi, and Andrew Zisserman.
\newblock Deep inside convolutional networks: Visualising image classification models and saliency maps.
\newblock In {\em International Conference on Learning Representations (ICLR) Workshop}, 2014.

\bibitem{zeiler2014deconvnet}
Matthew~D Zeiler and Rob Fergus.
\newblock Visualizing and understanding convolutional networks.
\newblock In {\em European Conference on Computer Vision (ECCV)}, pages 818--833. Springer, 2014.

\bibitem{springenberg2014guidedbp}
Jost~Tobias Springenberg, Alexey Dosovitskiy, Thomas Brox, and Martin Riedmiller.
\newblock Striving for simplicity: The all convolutional net.
\newblock In {\em International Conference on Learning Representations (ICLR) Workshop}, 2015.

\bibitem{smilkov2017smoothgrad}
Daniel Smilkov, Nikhil Thorat, Been Kim, Fernanda Vi{\'e}gas, and Martin Wattenberg.
\newblock Smoothgrad: removing noise by adding noise.
\newblock In {\em Workshop on Visualization for Deep Learning, International Conference on Machine Learning (ICML)}, 2017.

\bibitem{sundararajan2017ig}
Mukund Sundararajan, Ankur Taly, and Qiqi Yan.
\newblock Axiomatic attribution for deep networks.
\newblock In {\em Proceedings of the 34th International Conference on Machine Learning (ICML)}, pages 3319--3328, 2017.

\bibitem{zhou2016cam}
Bolei Zhou, Aditya Khosla, Agata Lapedriza, Aude Oliva, and Antonio Torralba.
\newblock Learning deep features for discriminative localization.
\newblock In {\em Proceedings of the IEEE Conference on Computer Vision and Pattern Recognition (CVPR)}, pages 2921--2929, 2016.

\bibitem{selvaraju2017gradcam}
Ramprasaath~R Selvaraju, Michael Cogswell, Abhishek Das, Ramakrishna Vedantam, Devi Parikh, and Dhruv Batra.
\newblock Grad-cam: Visual explanations from deep networks via gradient-based localization.
\newblock In {\em Proceedings of the IEEE International Conference on Computer Vision (ICCV)}, pages 618--626, 2017.

\bibitem{chattopadhay2018grad}
Aditya Chattopadhay, Anirban Sarkar, Prantik Howlader, and Vineeth~N Balasubramanian.
\newblock Grad-cam++: Generalized gradient-based visual explanations for deep convolutional networks.
\newblock In {\em 2018 IEEE winter conference on applications of computer vision (WACV)}, pages 839--847. IEEE, 2018.

\bibitem{jiang2021layercam}
Peng-Tao Jiang, Chang-Bin Zhang, Qibin Hou, Ming-Ming Cheng, and Yunchao Wei.
\newblock Layercam: Exploring hierarchical class activation maps for localization.
\newblock {\em IEEE transactions on image processing}, 30:5875--5888, 2021.

\bibitem{dosovitskiy2020image}
Alexey Dosovitskiy, Lucas Beyer, Alexander Kolesnikov, Dirk Weissenborn, Xiaohua Zhai, Thomas Unterthiner, Mostafa Dehghani, Matthias Minderer, Georg Heigold, Sylvain Gelly, et~al.
\newblock An image is worth 16x16 words: Transformers for image recognition at scale.
\newblock {\em arXiv preprint arXiv:2010.11929}, 2020.

\bibitem{abnar2020quantifying}
Samira Abnar and Willem Zuidema.
\newblock Quantifying attention flow in transformers.
\newblock {\em arXiv preprint arXiv:2005.00928}, 2020.

\bibitem{chefer2021transformer}
Hila Chefer, Shir Gur, and Lior Wolf.
\newblock Transformer interpretability beyond attention visualization.
\newblock In {\em Proceedings of the IEEE/CVF conference on computer vision and pattern recognition}, pages 782--791, 2021.

\bibitem{schlemper2023glaucoma}
Joana Schlemper and et~al.
\newblock In-depth evaluation of saliency maps for interpreting cnn decisions in the diagnosis of glaucoma.
\newblock {\em Translational Vision Science \& Technology}, 2023.

\bibitem{arun2020assessing}
Natekar~A. Arun, S.~Jain, and et~al.
\newblock Assessing the (un)trustworthiness of saliency maps for localizing abnormalities in medical imaging.
\newblock {\em medRxiv}, 2020.

\bibitem{kim2022pneumothorax}
Taehoon Kim and et~al.
\newblock Attention-based saliency maps improve interpretability of pneumothorax classification.
\newblock {\em Radiology: Artificial Intelligence}, 4(5), 2022.

\bibitem{zhang2023covid}
Wei Zhang and et~al.
\newblock Machine-learning-enabled diagnostics with improved visualization of disease lesions in chest x-ray images.
\newblock {\em Diagnostics}, 14(16):1699, 2023.

\bibitem{samek2017evaluating}
Wojciech Samek, Gr{\'e}goire Montavon, Andrea Vedaldi, Lars~Kai Hansen, and Klaus-Robert M{\"u}ller.
\newblock Evaluating the visualization of what a deep neural network has learned.
\newblock {\em IEEE Transactions on Neural Networks and Learning Systems}, 28(11):2660--2673, 2017.

\bibitem{fong2017meaningful}
Ruth~C. Fong and Andrea Vedaldi.
\newblock Interpretable explanations of black boxes by meaningful perturbation.
\newblock In {\em Proceedings of the IEEE International Conference on Computer Vision (ICCV)}, pages 3429--3437, 2017.

\bibitem{petsiuk2018rise}
Vitali Petsiuk, Abir Das, and Kate Saenko.
\newblock Rise: Randomized input sampling for explanation of black-box models.
\newblock In {\em British Machine Vision Conference (BMVC)}, 2018.

\bibitem{adebayo2018sanity}
Julius Adebayo, Justin Gilmer, Ian Muelly, Ian Goodfellow, Moritz Hardt, and Been Kim.
\newblock Sanity checks for saliency maps.
\newblock In {\em Advances in Neural Information Processing Systems (NeurIPS)}, volume~31, 2018.

\bibitem{koh2020concept}
Pang~Wei Koh, Thao Nguyen, Yew~Siang Tang, Stephen Mussmann, Everett Pierson, Been Kim, and Percy Liang.
\newblock Concept bottleneck models.
\newblock In {\em Proceedings of the 37th International Conference on Machine Learning (ICML)}, pages 5338--5348, 2020.

\bibitem{margeloiu2021cbmintended}
Andrei Margeloiu, Luca Costabello, Mateja Jamnik, Pietro Li{\`o}, and Adrian Weller.
\newblock Do concept bottleneck models learn as intended?
\newblock In {\em Neural Information Processing Systems (NeurIPS) Workshop on Explainable AI}, 2021.

\bibitem{oikarinen2023label}
Tuomas Oikarinen, Subhankar Adiraju, Rishika Rudin, and Christoph~H Lampert.
\newblock Label-free concept bottleneck models.
\newblock In {\em The Eleventh International Conference on Learning Representations}, 2023.

\bibitem{yuksekgonul2022posthoc}
Mert Yuksekgonul, Anshul Raghunathan, Dan Jurafsky, and Percy Liang.
\newblock Post-hoc concept bottleneck models.
\newblock In {\em Proceedings of the 2022 AAAI/ACM Conference on AI, Ethics, and Society}, pages 835--845, 2022.

\bibitem{marconato2023probabilistic}
Emanuele Marconato, Antonio Perri, Francesco Pozzobon, Marco Teso, and Andrea Passerini.
\newblock Probabilistic concept bottleneck models.
\newblock In {\em The Eleventh International Conference on Learning Representations}, 2023.

\bibitem{lampert2020concept}
Christoph~H Lampert.
\newblock Explaining decisions of deep neural networks by concept bottleneck models.
\newblock In {\em Proceedings of the German Conference on Artificial Intelligence (KI)}, 2020.

\bibitem{yuksekgonul2023posthoc}
Mert Yuksekgonul and et~al.
\newblock Beyond saliency: Post-hoc concept bottleneck models with language explanations.
\newblock {\em Nature Machine Intelligence}, 2023.

\bibitem{aysel2025conceptmetrics}
A.~Aysel and Others.
\newblock Concept-based explainable artificial intelligence: Metrics and benchmarks.
\newblock {\em arXiv preprint arXiv:2503.06873}, 2025.

\bibitem{chen2019looks}
Chaofan Chen, Oscar Li, Daniel Tao, Alina Barnett, Cynthia Rudin, and Jonathan~K Su.
\newblock This looks like that: deep learning for interpretable image recognition.
\newblock {\em Advances in neural information processing systems}, 32, 2019.

\bibitem{prototree2021}
Meike Nauta, Christin Seifert, and Markus Post.
\newblock Neural prototype trees for interpretable fine-grained image recognition.
\newblock In {\em Proceedings of the IEEE/CVF conference on computer vision and pattern recognition}, pages 14933--14943, 2021.

\bibitem{protopool2022}
Meike Nauta, Maurice van Bree, and Christin Seifert.
\newblock Interpretable image classification with differentiable prototypes assignment.
\newblock In {\em European Conference on Computer Vision}, pages 245--261. Springer, 2022.

\bibitem{pipnet2023}
Meike Nauta, Elise Buksan, and Christin Seifert.
\newblock Pip-net: Patch-based intuitive prototypes for interpretable image classification.
\newblock In {\em Proceedings of the IEEE/CVF Conference on Computer Vision and Pattern Recognition (CVPR)}, pages 11718--11728, 2023.

\bibitem{kim2018tcav}
Been Kim, Martin Wattenberg, Justin Gilmer, Carrie Cai, James Wexler, Fernanda Viegas, et~al.
\newblock Interpretability beyond feature attribution: Quantitative testing with concept activation vectors (tcav).
\newblock In {\em International conference on machine learning}, pages 2668--2677. PMLR, 2018.

\end{thebibliography}
\end{document}